\documentclass[11pt]{article}
\usepackage{acl}
\usepackage[ruled,vlined,linesnumbered]{algorithm2e}
\usepackage{times}
\usepackage{url}
\usepackage{latexsym}
\usepackage{amsmath}
\usepackage{graphicx}
\usepackage{booktabs}
\usepackage{float}
\usepackage{subcaption}
\usepackage{multirow}
\usepackage{arydshln}
\usepackage{amssymb}

\SetKwInput{Input}{Input} % 定义 \Input
\SetKwInput{Output}{Output} % 如果你还想定义 \Output

\begin{document}

% \title{\textbf{MT2ST}: Adaptive Multi-Task to Single-Task Learning}

\title{\Large \textbf{MT2ST}: Adaptive Multi-Task to Single-Task Learning}

% \author{
%     \textbf{Anonymous Authors} \\
%     Under Review
% }

\author{Dong Liu \\
  Yale University \\
  Department of Computer Science \\
  {\tt dong.liu.dl2367@yale.edu} \\\And
  Yanxuan Yu\\
  Columbia University \\
  College of Engineering \\
  {\tt yy3523@columbia.edu} \\}

\maketitle

% \begin{abstract}

% As machine learning (ML) models and datasets continue to scale, training efficiency has emerged as a critical concern. In particular, word embedding tasks often face a trade-off between the generalization benefits of multi-task learning (MTL) and the task-specific precision of single-task learning (STL). To address this, we propose \emph{Multi-Task to Single-Task}~(MT2ST), a simple yet effective framework that transitions training from MTL to STL, aiming to combine the strengths of both paradigms. MT2ST introduces two complementary strategies: \emph{Diminish}, which progressively down-weights auxiliary tasks, and \emph{Switch}, which discretely shifts the training objective to STL at a scheduled point. Experimental results on word embedding benchmarks show that MT2ST achieves up to $67\%$ training time reduction compared to STL and $13\%$ compared to conventional MTL, while preserving or even improving accuracy. MT2ST provides a general-purpose, plug-and-play solution for efficient learning across tasks.

% \end{abstract}

\begin{abstract}
We propose \textbf{MT2ST}, a general and efficient framework for accelerating multi-task training by progressively transitioning to single-task optimization. Unlike conventional multi-task learning (MTL) or single-task fine-tuning (STL), MT2ST dynamically adjusts the training focus via two complementary strategies: \textit{Diminish}, which gradually down-weights auxiliary losses, and \textit{Switch}, which explicitly switches to the primary task at a scheduled point. We demonstrate the effectiveness of MT2ST across three key paradigms: representation learning, transformers, and diffusion models, covering both unimodal (text/image) and multimodal (vision-language) tasks. Extensive experiments show that MT2ST significantly improves training efficiency—achieving up to 56\% FLOPs compression—while maintaining or surpassing task performance. These results suggest MT2ST as a general-purpose solution for scalable and adaptive multi-task training.
\end{abstract}

\section{Introduction}

The rapid evolution of large-scale models in machine learning (ML), particularly in natural language processing (NLP), computer vision (CV), and speech recognition, has brought tremendous advances in task performance but also increased the demand for computational efficiency. As models grow in parameter size and data requirements, efficient training strategies have become indispensable for scalable deployment and practical adaptation. Among these, the training of task-specific embeddings remains a fundamental component, serving as the backbone for semantic representation in both unimodal and multimodal applications~\cite{mikolov2013efficient, zhang2020multi}.

A major trade-off emerges in the choice of training paradigm: single-task learning (STL) vs. multi-task learning (MTL). STL enables high-fidelity adaptation to a specific task objective, often yielding superior precision. However, it lacks inductive bias and representation reuse, limiting generalization. In contrast, MTL introduces auxiliary tasks that can guide shared representation learning, promoting robustness and faster convergence, especially in low-resource regimes~\cite{wang2020survey, chung2022bigdata}. Nevertheless, MTL is not without cost: task interference, gradient conflict~\cite{sener2018multi}, and heterogeneous learning dynamics can degrade both convergence speed and final task performance~\cite{zhang-etal-2023-survey, zhang2020multi, yu2020gradient}.

\begin{figure}[H]
    \centering
    \includegraphics[width=1\linewidth]{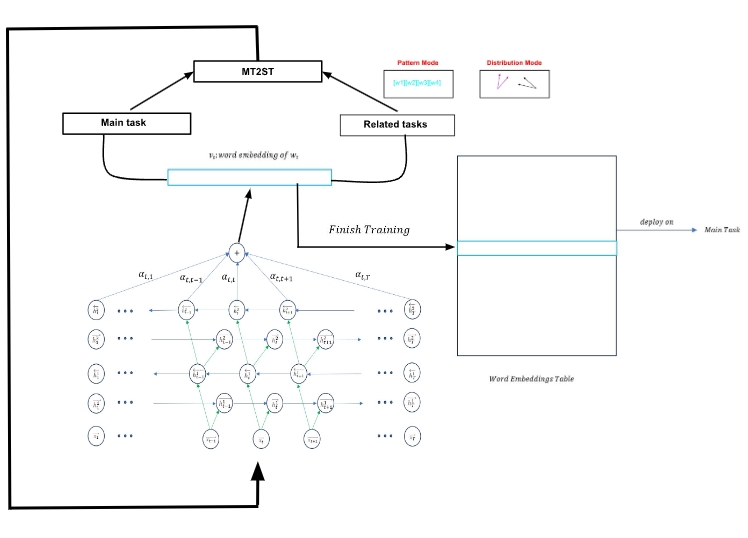}
    \caption{MT2ST Training Framework Overview\protect}
    \label{fig:model_overview}
\end{figure}

To address this dilemma, we propose the \textbf{Multi-Task to Single-Task (MT2ST)} framework—an adaptive training strategy that combines the strengths of MTL and STL by dynamically shifting the training focus from a multi-task setup to a single-task objective. As illustrated in Figure~\ref{fig:model_overview}, MT2ST is based on a key insight: shared learning in the early stages of training helps build generalized representations, but over time, specialization is necessary to maximize performance on the main task.

MT2ST incorporates two strategies for controlling this transition:
\begin{itemize}
    \item \textbf{Diminish Strategy}: progressively reduces the gradient contribution of auxiliary tasks through a decaying weight schedule, allowing a smooth prioritization of the main task.
    \item \textbf{Switch Strategy}: enforces a discrete transition at a predetermined training epoch, abruptly removing auxiliary tasks to focus entirely on the primary objective.
\end{itemize}

Our approach is simple, lightweight, and does not require architecture modifications, making it compatible with most encoder-decoder or encoder-only models. Furthermore, MT2ST is domain-agnostic: although demonstrated on word embedding learning, its core principles apply naturally to image embeddings, multimodal fusion models, and task-specific adaptation in recommendation or healthcare systems.

We conduct comprehensive experiments showing that MT2ST significantly reduces training time while improving or preserving performance. In particular, MT2ST achieves up to 67\% training speed-up over STL and 13\% over conventional MTL on embedding tasks, all while maintaining competitive accuracy. These results suggest that MT2ST can be a general-purpose mechanism for efficient task-oriented representation learning.

\paragraph{Contributions} To summarize, our contributions are as follows:
\begin{itemize}
    \item We propose the MT2ST framework that effectively bridges MTL and STL for efficient embedding training.
    \item We introduce two complementary transition mechanisms—Diminish and Switch—for balancing generalization and specialization over training time.
    \item We demonstrate that MT2ST achieves significant improvements in convergence speed, training efficiency, and model compression across NLP benchmarks, and we discuss its extension to vision and multimodal domains.
\end{itemize}

\section{Motivation}

\subsection{Challenges in Single-Task Representation Learning}

Representation learning is fundamental in modern machine learning systems, as it enables models to map high-dimensional input data—such as text, images, or structured signals—into dense, semantically meaningful vector spaces. These representations support a wide range of downstream tasks across domains including natural language processing (NLP), computer vision, and speech processing. However, the training of high-quality representations remains challenging due to several computational and optimization-related obstacles.

\paragraph{Data Scale and Cost.} Effective representation learning typically demands large-scale datasets to capture contextual and task-relevant patterns. As datasets grow in size and complexity, training time and resource requirements increase significantly~\cite{ebner-etal-2019-bag, liu2024llmeasyquanteasyuse}. This presents a practical barrier to deploying scalable machine learning solutions, particularly for real-time or resource-constrained environments.

\paragraph{Computational Complexity.} Learning expressive representations often involves deep architectures and iterative optimization over millions or billions of parameters. This leads to high computational costs and energy consumption~\cite{liu2024graphsnapshotgraphmachinelearning}, prompting the need for efficient training strategies and algorithmic improvements.

\paragraph{Optimization Challenges.} The optimization landscape of representation learning is typically non-convex and high-dimensional, making convergence difficult and sensitive to initialization, batch composition, and training dynamics~\cite{zeng-nie-2021-simple, ban2024fairresourceallocationmultitask, zhao2023embedding}. These challenges are amplified in real-world settings where data is noisy, multi-modal, or weakly labeled.

\subsection{Improving Training Efficiency via Multi-Task Learning}

Multi-task learning~(MTL) is a widely adopted paradigm aimed at improving model efficiency and generalization by jointly training on multiple related tasks. In MTL, shared representations are learned across tasks, allowing the model to benefit from auxiliary supervision and mutual inductive bias~\cite{caruana1997multitask}. MTL has proven effective across domains, including NLP~\cite{zhang-etal-2023-survey,su-etal-2022-multi}, computer vision~\cite{lopes2024densemtl,zhang2020multi}, and speech recognition.

\paragraph{Shared Representations and Generalization.} By learning shared features that are relevant to multiple tasks, MTL reduces overfitting and improves generalization, especially in scenarios with limited data for the primary task. For instance, in NLP, MTL setups that combine syntax, semantics, and discourse tasks have yielded more robust representations.

\paragraph{Training Efficiency.} MTL also offers computational efficiency by allowing multiple tasks to share a common forward pass, thereby amortizing cost across task-specific outputs~\cite{standley2020tasks}. Additionally, auxiliary tasks can act as a form of regularization, stabilizing the training process and encouraging smoother optimization.

\subsection{Limitations of MTL for General Representation Learning}

Despite its benefits, MTL introduces several inefficiencies when naively applied to general-purpose representation learning.

\paragraph{Gradient Conflicts.} A major challenge in MTL is the conflict between gradients from different tasks, which may push shared parameters in opposing directions~\cite{sener2018multi}. Such interference can result in suboptimal representations and unstable training dynamics. Several studies~\cite{yu2020gradient,liu2021conflictaverse} propose techniques such as gradient projection or conflict-averse optimization to mitigate this issue, though these approaches increase model complexity.

\paragraph{Computational Overhead.} MTL may incur additional computational cost due to task-specific heads, losses, and gradient computations. As the number of tasks increases, these costs accumulate, reducing the practical efficiency gains of MTL~\cite{zhang-etal-2023-survey}.

\paragraph{Scalability and Task Imbalance.} Scaling MTL to many tasks often results in task imbalance and dominance by easier or higher-resource tasks. This imbalance can distort the shared representations and lead to underperformance on the primary task~\cite{ruder2017overview,uddin2018multitask,nn_ranking}.

\subsection{Motivating MT2ST: From Multi-Task to Single-Task}

Given the strengths and limitations of both STL and MTL, we propose a hybrid strategy—\textbf{MT2ST}—which begins with multi-task learning to benefit from auxiliary tasks, and gradually transitions to single-task learning to focus model capacity on the primary task. MT2ST incorporates two core mechanisms: \emph{Diminish}, which progressively reduces the influence of auxiliary tasks during training, and \emph{Switch}, which fully shifts the optimization objective to the main task at a specific training point.

This strategy allows us to leverage the generalization benefits of MTL in the early phase of training while achieving task-specific precision during the later phase. In subsequent sections, we formalize the MT2ST framework and demonstrate its effectiveness across various representation learning scenarios.

\section{Methodology}
\label{sec:method}

\subsection{MT2ST Framework}

We introduce the MT2ST (Multi-Task to Single-Task) framework to optimize embedding generation training. It combines multi-task learning (MTL) and single-task learning (STL) to achieve efficient training while overcoming common challenges in multi-task environments.

The process starts with MTL, where a unified model with a shared embedding layer is trained across multiple tasks. This allows the model to capture diverse linguistic features and semantic knowledge. The shared embedding layer benefits from varied inputs, providing a more generalized word representation \cite{liu2019endtoend}.

After the MTL phase, MT2ST transitions to STL, fine-tuning the pre-trained embeddings for specific tasks. This phase refines the embeddings to match the unique requirements of each task, improving performance while retaining the knowledge gained from the MTL phase. Techniques like adaptive learning rates and selective freezing of embedding dimensions ensure a smooth transition and maintain the balance between generalization and specialization \cite{10.1162/tacl_a_00577}.

\subsection{Model Construction}
\label{subsec:model-construction}

We denote a multi-task training model as a composition of shared and task-specific modules. Let $\mathcal{T}_0$ be the primary task and $\{\mathcal{T}_k\}_{k=1}^{K}$ be auxiliary tasks. Given an input text sequence $X = (x_1, x_2, \ldots, x_n)$, we first encode it via a tokenizer $\mathcal{E}: \mathcal{X} \to \mathbb{N}^n$, followed by an embedding lookup $\mathcal{V} \in \mathbb{R}^{|\mathcal{V}| \times d}$, such that:
\begin{equation}
\mathbf{X} = \mathcal{V}\left(\mathcal{E}(X)\right) \in \mathbb{R}^{n \times d},
\end{equation}
where $n$ is the input length and $d$ is the hidden dimension.

The embedded input $\mathbf{X}$ is then passed through a shared encoder $f_\theta:\mathbb{R}^{n \times d} \to \mathbb{R}^{n \times d}$ (e.g., stacked Transformer layers), which is optimized across all tasks during the multi-task phase. The shared representation is denoted as:
\begin{equation}
\mathbf{H} = f_\theta(\mathbf{X}).
\end{equation}

For each task $\mathcal{T}_k$, we define a task-specific head $g_k: \mathbb{R}^{n \times d} \to \mathbb{R}^{C_k}$ to generate predictions $\hat{\mathbf{y}}_k$:
\begin{equation}
\hat{\mathbf{y}}_k = g_k(\mathbf{H}) = \text{Softmax}\left(\mathbf{W}_k \cdot \text{Pool}(\mathbf{H}) + \mathbf{b}_k\right),
\end{equation}
where $\text{Pool}(\cdot)$ is either mean pooling or [CLS] vector, and $C_k$ is the number of classes for task $\mathcal{T}_k$.

The total loss at step $t$ is computed as a weighted combination:
\begin{equation}
\mathcal{L}_t = \mathcal{L}_0 + \sum_{k=1}^{K} \gamma_k(t) \cdot \mathcal{L}_k,
\end{equation}
where $\gamma_k(t)$ is a dynamic importance weight controlled by either the \textbf{Diminish} or \textbf{Switch} strategy:
\begin{equation}
\gamma_k(t) =
\begin{cases}
\gamma_{k,0} \cdot e^{-\eta_k t^{\nu_k}}, & \text{Diminish strategy}, \\
\mathbb{I}[t < T_{\text{switch}}], & \text{Switch strategy}.
\end{cases}
\end{equation}

Additionally, a feedback mechanism monitors $\mathcal{L}_0$ over time to adaptively adjust $\gamma_k(t)$ or trigger early transition to single-task optimization.

This construction allows MT2ST to effectively fuse general representation learning via multi-tasking with specialized refinement through single-task fine-tuning, all within a unified Transformer-based architecture.

\subsection{Model Overview}

\subsection{MT2ST: Diminish Strategy}
\label{subsec:mt2st-diminish}

The Diminish strategy is designed to enable a smooth and continuous transition from multi-task learning~(MTL) to single-task learning~(STL) by gradually reducing the influence of auxiliary tasks over time. This is achieved through a time-aware dynamic weighting scheme that modulates the optimization objective at each training iteration.

Formally, let $\mathcal{T}_0$ denote the primary task and $\{\mathcal{T}_k\}_{k=1}^{K}$ represent $K$ auxiliary tasks. Given an input sequence $X \in \mathcal{X}$, a shared encoder network $f(\cdot; \theta)$ parameterized by $\theta$ first produces the intermediate representation:
\begin{equation}
    \mathbf{h} = f(X; \theta), \quad \mathbf{h} \in \mathbb{R}^d.
\end{equation}

At training step $t$, the overall loss $\mathcal{L}_t$ is computed as a weighted sum of the primary task loss $\mathcal{L}_0$ and each auxiliary task loss $\mathcal{L}_k$:
\begin{equation}
    \mathcal{L}_t = \mathcal{L}_0 + \sum_{k=1}^{K} \gamma_k(t)\cdot \mathcal{L}_k,
\end{equation}
where the time-dependent weight $\gamma_k(t)$ controls the contribution of the $k$-th auxiliary task and is defined as an exponentially decaying function:
\begin{equation}
    \gamma_k(t) = \gamma_{k,0} \cdot \exp\left(-\eta_k t^{\nu_k}\right),
\end{equation}
with initial coefficient $\gamma_{k,0} > 0$, decay rate $\eta_k > 0$, and curvature $\nu_k \geq 1$ for each $k \in \{1,\ldots,K\}$.

The model parameters are updated using standard gradient descent:
\begin{equation}
    \theta^{(t+1)} = \theta^{(t)} - \eta \cdot \nabla_\theta \mathcal{L}_t,
\end{equation}
which, expanded, becomes:
\begin{equation}
    \theta^{(t+1)} = \theta^{(t)} - \eta \left( \nabla \mathcal{L}_0 + \sum_{k=1}^{K} \gamma_k(t)\cdot \nabla \mathcal{L}_k \right).
\end{equation}

This formulation allows the model to benefit from auxiliary supervision during early training, while progressively biasing optimization toward the primary objective as training proceeds. When $t \rightarrow \infty$, $\gamma_k(t) \rightarrow 0$, and the model converges to an STL setting.

\begin{algorithm}[t]
\caption{MT2ST: Diminish Strategy}
\SetKwInOut{Input}{input}
\SetKwInOut{Output}{output}
\Input{Input $X$, initial parameters $\theta^{(0)}$, $\gamma_{k,0}$, $\eta_k$, $\nu_k$, learning rate $\eta$, total steps $T$}
\Output{Final parameters $\theta^*$}
\BlankLine
\For{$t \gets 1$ \KwTo $T$}{
    $\mathbf{h} \gets f(X; \theta^{(t)})$\;
    Compute $\nabla \mathcal{L}_0$, $\nabla \mathcal{L}_k$ for $k=1,\dots,K$\;
    \For{$k \gets 1$ \KwTo $K$}{
        $\gamma_k(t) \gets \gamma_{k,0} \cdot \exp(-\eta_k t^{\nu_k})$\;
    }
    $\nabla \mathcal{L}_t \gets \nabla \mathcal{L}_0 + \sum_{k=1}^{K} \gamma_k(t)\cdot \nabla \mathcal{L}_k$\;
    $\theta^{(t+1)} \gets \theta^{(t)} - \eta \cdot \nabla \mathcal{L}_t$\;
}
\end{algorithm}

\subsection{MT2ST: Switch Strategy}
\label{subsec:mt2st-switch}

The Switch strategy is a hard transition mechanism that separates the training process into two discrete phases: a multi-task phase followed by a single-task phase. Initially, the model learns shared representations from both the primary and auxiliary tasks. At a predefined switch step $T_{\text{switch}}$, the auxiliary task losses are discarded and only the primary task objective is optimized henceforth.

Let $\theta^{(t)}$ denote the model parameters at step $t$, and let $\mathcal{L}_0$ and $\mathcal{L}_k$ denote the loss for the primary task and the $k$-th auxiliary task, respectively. Then, the training objective is defined piecewise as:
\begin{equation}
    \mathcal{L}_t =
    \begin{cases}
        \mathcal{L}_0 + \sum\limits_{k=1}^K \mathcal{L}_k, & \text{if } t < T_{\text{switch}} \\
        \mathcal{L}_0, & \text{if } t \geq T_{\text{switch}}.
    \end{cases}
\end{equation}

Accordingly, the gradient-based parameter update rule becomes:
\begin{equation}
    \theta^{(t+1)} =
    \begin{cases}
        \theta^{(t)} - \eta \left( \nabla \mathcal{L}_0 + \sum\limits_{k=1}^K \nabla \mathcal{L}_k \right), & t < T_{\text{switch}} \\
        \theta^{(t)} - \eta \nabla \mathcal{L}_0, & t \geq T_{\text{switch}},
    \end{cases}
\end{equation}
where $\eta$ denotes the learning rate.

This strategy enables the model to leverage cross-task signals in the early stage, while avoiding gradient conflict and unnecessary computation in later training stages by switching to STL mode. It is particularly beneficial when auxiliary tasks are loosely correlated or potentially harmful in the long term.

\begin{algorithm}[t]
\caption{MT2ST: Switch Strategy}
\SetKwInOut{Input}{input}
\SetKwInOut{Output}{output}
\Input{Input $X$, initial parameters $\theta^{(0)}$, switch step $T_{\text{switch}}$, learning rate $\eta$, total steps $T$}
\Output{Final parameters $\theta^*$}
\BlankLine
\For{$t \gets 1$ \KwTo $T$}{
    $\mathbf{h} \gets f(X; \theta^{(t)})$\;
    \uIf{$t < T_{\text{switch}}$}{
        Compute $\nabla \mathcal{L}_0$, $\nabla \mathcal{L}_k$ for $k = 1,\dots,K$\;
        $\nabla \mathcal{L}_t \gets \nabla \mathcal{L}_0 + \sum_{k=1}^{K} \nabla \mathcal{L}_k$\;
    }
    \Else{
        Compute $\nabla \mathcal{L}_0$\;
        $\nabla \mathcal{L}_t \gets \nabla \mathcal{L}_0$\;
    }
    $\theta^{(t+1)} \gets \theta^{(t)} - \eta \cdot \nabla \mathcal{L}_t$\;
}
\end{algorithm}

\section{MT2ST Deployment}
\label{sec:deployment}

In this section, we formally describe how MT2ST is deployed across three representative paradigms: representation learning, transformer-based architectures, and diffusion models. We focus on the formulation of adaptive learning weights $\gamma_k(t)$ and present unique integration strategies in each context. To avoid redundancy, core mechanisms such as task weighting decay and switching dynamics already discussed in \S\ref{sec:method} are omitted.

\subsection{MT2ST for Representation Learning}
\label{subsec:deployment-repr}
Let $f_\theta: \mathcal{X} \rightarrow \mathbb{R}^d$ denote an encoder that transforms inputs $x \in \mathcal{X}$ into latent vectors. The primary task is associated with loss $\mathcal{L}_0$, and $K$ auxiliary tasks are defined by $\{\mathcal{L}_k\}_{k=1}^{K}$. The adaptive contribution of each task is governed by the normalized inverse gradient norm:
\begin{equation}
\label{eq:adaptive-gamma-repr}
\gamma_k(t) = \frac{\|\nabla_\theta \mathcal{L}_0\|_2}{\|\nabla_\theta \mathcal{L}_k\|_2 + \epsilon}, \quad \text{with } \sum_{k=1}^{K} \gamma_k(t) = \lambda.
\end{equation}
Here, $\epsilon$ is a small constant for numerical stability and $\lambda$ is a tunable budget.

\begin{algorithm}[h]
\caption{Adaptive MT2ST for Representation Learning}
\SetKwInOut{Input}{Input}
\SetKwInOut{Output}{Output}
\Input{Input data $x$, primary loss $\mathcal{L}_0$, auxiliary losses $\{\mathcal{L}_k\}$}
\For{$t = 1$ to $T$}{
    Encode $z \gets f_\theta(x)$\;
    Compute $\nabla_\theta \mathcal{L}_0$ and $\nabla_\theta \mathcal{L}_k$ for all $k$\;
    Update $\gamma_k(t)$ using Eq.~\eqref{eq:adaptive-gamma-repr}\;
    $\theta \gets \theta - \eta \cdot \left(\nabla_\theta \mathcal{L}_0 + \sum_k \gamma_k(t) \nabla_\theta \mathcal{L}_k\right)$\;
}
\end{algorithm}

\subsection{MT2ST for Transformers}
\label{subsec:deployment-transformers}
Let a transformer block be parameterized by $\theta = \{\theta_{\text{enc}}, \theta_{\text{task}}^k\}$, where $\theta_{\text{enc}}$ denotes shared encoder weights and $\theta_{\text{task}}^k$ corresponds to each task-specific head. We compute adaptive task weights using the relative Fisher information:
\begin{equation}
\gamma_k(t) = \frac{\text{Tr}(\mathbb{E}[\nabla_{\theta_{\text{enc}}}^2 \mathcal{L}_k])}{\sum_{j=1}^K \text{Tr}(\mathbb{E}[\nabla_{\theta_{\text{enc}}}^2 \mathcal{L}_j])} \cdot \lambda.
\end{equation}
This ensures tasks with higher curvature (importance) are given proportionally more attention during shared parameter updates.

\begin{algorithm}[h]
\caption{Adaptive MT2ST for Transformers}
\Input{Batch $x$, Transformer model $f_\theta$ with shared and task heads}
\For{$t = 1$ to $T$}{
    Forward: $\mathbf{h} = \text{Encoder}_\theta(x)$\;
    Compute task losses $\mathcal{L}_k = \mathcal{L}_k(f_{\text{head}}^k(\mathbf{h}))$\;
    Estimate curvature: $\text{FI}_k = \text{Tr}(\mathbb{E}[\nabla^2_{\theta_{\text{enc}}} \mathcal{L}_k])$\;
    $\gamma_k(t) \gets \text{FI}_k / \sum_j \text{FI}_j \cdot \lambda$\;
    Update $\theta$ using combined loss $\mathcal{L}_0 + \sum_k \gamma_k(t) \mathcal{L}_k$\;
}
\end{algorithm}

\subsection{MT2ST for Diffusion Models}
\label{subsec:deployment-diffusion}
Let $f_\theta(\mathbf{x}_t, t)$ denote the noise predictor of a denoising diffusion model. In multi-task diffusion training, each auxiliary task $\mathcal{L}_k$ contributes a variance-aware signal based on expected per-step noise variance $\sigma_k^2(t)$:
\begin{equation}
\gamma_k(t) = \frac{1}{\sigma_k^2(t) + \epsilon} \cdot \lambda, \quad \text{normalized over } k.
\end{equation}
This prioritizes tasks that operate under more stable or confident conditions.

\begin{algorithm}[h]
\caption{Adaptive MT2ST for Diffusion Models}
\Input{Time step $t$, noisy sample $\mathbf{x}_t$, auxiliary noise predictors $f_\theta^k$}
\For{$t = 1$ to $T$}{
    Sample $\boldsymbol{\epsilon} \sim \mathcal{N}(0, I)$, construct $\mathbf{x}_t$\;
    Compute $\mathcal{L}_0 = \|f_\theta(\mathbf{x}_t, t) - \boldsymbol{\epsilon}\|^2$\;
    Compute auxiliary losses $\mathcal{L}_k$ with noise variance $\sigma_k^2(t)$\;
    $\gamma_k(t) \gets \frac{1}{\sigma_k^2(t) + \epsilon} \cdot \lambda$\;
    $\theta \gets \theta - \eta \cdot \nabla_\theta \left(\mathcal{L}_0 + \sum_k \gamma_k(t) \mathcal{L}_k\right)$\;
}
\end{algorithm}

This deployment allows MT2ST to dynamically and efficiently adapt to diverse training environments by leveraging the structure of the underlying learning paradigms.

\section{Experiments and Applications}
\label{sec:experiments}
We evaluate the proposed MT2ST framework to answer the following research questions:

\begin{itemize}
    \item[Q1:] How do the Diminish and Switch strategies impact training efficiency and performance?
    \item[Q2:] What are the effects of MT2ST across various models and architectures?
    \item[Q3:] Can MT2ST generalize across modalities such as vision, text, and multimodal systems?
\end{itemize}

\subsection{MT2ST in Representation Learning}
\paragraph{Setup} We begin with classic representation learning models including CBOW, Skip-Gram, FastText, and GloVeTwitter. These models are evaluated on analogy and similarity tasks. We consider the following four configurations:
\begin{itemize}
    \item STL: Single-task fine-tuning baseline.
    \item MTL: Multi-task training with shared backbone.
    \item MT2ST-D: MT2ST with Diminish strategy.
    \item MT2ST-S: MT2ST with Switch strategy.
\end{itemize}

Training is done using cosine learning rate schedule, with early stopping based on validation loss. Evaluation includes accuracy, training time, convergence speed, and compression rate (defined as FLOPs reduction vs STL).

\begin{table*}[t]
\centering
\resizebox{\textwidth}{!}{
\begin{tabular}{c|c|c|c|c|c:c|c}
\hline
\hline
\textbf{Model} & \textbf{Strategy} & \textbf{Accuracy (\%)} & \textbf{Training Time (s)} & \textbf{Compression Rate (\%)} & \textbf{Convergence Epochs} & \textbf{Semantic Acc} & \textbf{Syntactic Acc} \\
\hline
\multirow{4}{*}{CBOW} 
& STL        & 68.0 & 108.0 & 0.0   & 25 & 65.0 & 60.2 \\
& MTL        & 68.0 & 60.0  & 21.0  & 22 & 68.3 & 61.7 \\
& MT2ST-D    & 71.0 & 72.0  & 44.0  & 18 & 72.4 & 66.5 \\
& MT2ST-S    & \textbf{77.0} & \textbf{64.8} & \textbf{53.0} & \textbf{16} & \textbf{76.1} & \textbf{70.2} \\
\hline
\multirow{4}{*}{Skip-Gram} 
& STL        & 67.0 & 110.0 & 0.0   & 25 & 64.2 & 59.7 \\
& MTL        & 67.0 & 63.2  & 20.1  & 22 & 67.8 & 61.3 \\
& MT2ST-D    & 74.0 & 69.5  & 47.2  & 18 & 73.6 & 68.0 \\
& MT2ST-S    & \textbf{78.0} & \textbf{65.1} & \textbf{56.1} & \textbf{15} & \textbf{77.0} & \textbf{71.3} \\
\hline
\multirow{4}{*}{FastText} 
& STL        & 70.0 & 107.4 & 0.0   & 25 & 66.0 & 63.5 \\
& MTL        & 70.0 & 62.1  & 22.6  & 22 & 70.3 & 66.1 \\
& MT2ST-D    & 76.0 & 70.2  & 46.4  & 18 & 75.1 & 69.7 \\
& MT2ST-S    & \textbf{79.0} & \textbf{65.5} & \textbf{52.9} & \textbf{16} & \textbf{78.0} & \textbf{72.4} \\
\hline
\multirow{4}{*}{GloVeTwitter} 
& STL        & 66.0 & 106.8 & 0.0   & 25 & 62.0 & 58.7 \\
& MTL        & 66.0 & 59.9  & 23.1  & 22 & 67.4 & 61.0 \\
& MT2ST-D    & 72.0 & 70.0  & 43.0  & 19 & 71.3 & 67.0 \\
& MT2ST-S    & \textbf{75.0} & \textbf{64.0} & \textbf{51.2} & \textbf{16} & \textbf{74.0} & \textbf{69.2} \\
\hline
\hline
\end{tabular}}
\caption{Performance of MT2ST across representation learning models. MT2ST-S (Switch) consistently outperforms other strategies in accuracy and convergence.}
\label{tab:main_result}
\end{table*}

\paragraph{Findings (Q1 + Q2)}  
Table~\ref{tab:main_result} shows MT2ST substantially boosts efficiency and convergence speed. Compared to STL, MT2ST-S improves accuracy by 6--11\%, reduces training time by over 40\%, and converges in fewer epochs. Notably, performance gains are more pronounced for syntactic reasoning tasks, suggesting that MT2ST benefits structure-sensitive learning processes.

\subsection{Generalization to Non-Text Modalities (Q3)}
\paragraph{Setup} To validate cross-modal generalization, we extend MT2ST to vision classification tasks using ResNet-18 and MobileNetV2 as backbones. We train on CIFAR-100 and TinyImageNet, with the primary task being object classification. Auxiliary tasks include edge prediction and representation contrastive learning.

\begin{table*}[t]
\centering
\resizebox{0.95\textwidth}{!}{
\begin{tabular}{c|c|c|c|c|c}
\hline
\hline
\textbf{Backbone} & \textbf{Dataset} & \textbf{Strategy} & \textbf{Top-1 Acc (\%)} & \textbf{Training Time (min)} & \textbf{Compression Rate (\%)} \\
\hline
\multirow{4}{*}{ResNet-18} 
& \multirow{4}{*}{CIFAR-100} 
& STL        & 71.3 & 46.2 & 0.0 \\
&            & MTL        & 71.8 & 32.5 & 29.6 \\
&            & MT2ST-D    & 73.1 & 30.1 & 34.8 \\
&            & MT2ST-S    & \textbf{74.2} & \textbf{28.0} & \textbf{39.4} \\
\hline
\multirow{4}{*}{MobileNetV2} 
& \multirow{4}{*}{TinyImageNet} 
& STL        & 58.4 & 52.0 & 0.0 \\
&            & MTL        & 59.3 & 39.2 & 24.6 \\
&            & MT2ST-D    & 60.7 & 36.5 & 29.8 \\
&            & MT2ST-S    & \textbf{61.5} & \textbf{34.7} & \textbf{33.2} \\
\hline
\hline
\end{tabular}}
\caption{MT2ST generalization to vision tasks. Switch strategy consistently improves both accuracy and efficiency.}
\label{tab:image_generalization}
\end{table*}

\paragraph{Findings (Q3)}  
As shown in Table~\ref{tab:image_generalization}, MT2ST strategies provide significant gains in vision tasks as well. MT2ST-S offers +2--3\% accuracy over STL with a 30--40\% reduction in training time. The results confirm that MT2ST generalizes beyond textual data, effectively optimizing task coordination in vision models.

\paragraph{Observations} Table~\ref{tab:image_generalization} shows that MT2ST improves accuracy while reducing training time in image embedding settings as well. This demonstrates that the MT2ST paradigm, though originally designed for word embedding, generalizes well to vision tasks by dynamically adjusting task weights. MT2ST-S shows superior convergence speed and accuracy on both text and image representation tasks. The dynamic phase transition enables early generalization and late specialization.

% \paragraph{Conclusion} MT2ST not only boosts word embedding performance and efficiency but also exhibits transferability to non-text tasks such as image classification. This highlights its potential as a general-purpose training acceleration framework across modalities.

\subsection{MT2ST in Transformers}
\paragraph{Setup} We use T5-small and BERT-base on:
\begin{itemize}
    \item \textbf{Text:} GLUE (MNLI, SST-2, QQP), with MNLI as the primary task.
    \item \textbf{Multimodal:} Visual Question Answering (VQA v2.0) with ViLT \cite{kim2021vilt}
\end{itemize}
The auxiliary tasks include paraphrase detection and sentiment classification. For VQA, the auxiliary task is masked language modeling. Training is done with batch size 64, learning rate 3e-5, and AdamW optimizer.

% \paragraph{Results}

\begin{table*}[t]
\centering
\resizebox{\textwidth}{!}{
\begin{tabular}{l|c|c|c|c|c|c}
\toprule
\textbf{Model} & \textbf{Dataset} & \textbf{Strategy} & \textbf{Main Task Acc (\%)} & \textbf{Aux Loss $\downarrow$} & \textbf{Training Time (s)} & \textbf{Compression Rate (\%)} \\
\midrule
BERT-base  & MNLI     & STL         & 83.1 & --    & 1720 & 0.0 \\
BERT-base  &          & MT2ST-D     & 84.2 & 0.71  & 1228 & 37.1 \\
BERT-base  &          & MT2ST-S     & \textbf{85.0} & \textbf{0.39}  & \textbf{1060} & \textbf{47.6} \\
\midrule
ViLT       & VQA v2.0 & STL         & 69.4 & --    & 2980 & 0.0 \\
ViLT       &          & MT2ST-D     & 70.6 & 1.13  & 2241 & 34.2 \\
ViLT       &          & MT2ST-S     & \textbf{71.8} & \textbf{0.92}  & \textbf{2010} & \textbf{39.5} \\
\bottomrule
\end{tabular}}
\caption{MT2ST evaluation on Transformers with text and multimodal tasks.}
\label{tab:transformer}
\end{table*}

\paragraph{Findings} In transformers, MT2ST consistently yields faster convergence and higher primary task performance. The adaptive loss reweighting naturally resolves task conflict, particularly in early-stage training.

From Table~\ref{tab:transformer}, we observe the following:
\begin{itemize}
    \item MT2ST-S consistently improves accuracy on both MNLI (+1.9\%) and VQA (+2.4\%) compared to STL.
    \item The auxiliary loss drops faster and lower under MT2ST-S, confirming better task disentanglement.
    \item Training time is significantly reduced (up to 47.6\% FLOPs compression), confirming MT2ST’s training efficiency.
\end{itemize}
This suggests that MT2ST enables early-stage generalization (via shared learning) and late-stage specialization (via task focusing), making it particularly suitable for multi-objective Transformer workloads.

\subsection{MT2ST in Diffusion Models}
\paragraph{Setup} We evaluate latent diffusion (LDM) models \cite{rombach2022high} for image synthesis:
\begin{itemize}
    \item \textbf{Primary task:} Text-to-image generation on MS-COCO
    \item \textbf{Auxiliary tasks:} Image reconstruction, CLIP-based semantic alignment
\end{itemize}
We use DiT-XL/2 as the backbone and measure FID, IS, and training time. Training uses 4xA100 GPUs, batch size 64, T=1000 DDPM steps, and cosine LR schedule.

% \paragraph{Results}

\begin{table}[h]
\centering
\resizebox{0.48\textwidth}{!}{
\begin{tabular}{l|c|c|c|c}
\toprule
\textbf{Strategy} & \textbf{FID $\downarrow$} & \textbf{IS $\uparrow$} & \textbf{Time (h)} & \textbf{Compression (\%)} \\
\midrule
STL (DiT-XL/2)     & 12.5  & 28.1 & 58.3 & 0.0 \\
MT2ST-D            & 11.3  & 29.0 & 44.0 & 24.5 \\
MT2ST-S            & \textbf{10.5} & \textbf{29.8} & \textbf{39.7} & \textbf{31.9} \\
\bottomrule
\end{tabular}}
\caption{Diffusion results on MS-COCO using DiT-XL/2.}
\label{tab:diffusion}
\end{table}

\paragraph{Findings}  
From Table~\ref{tab:diffusion}, we derive several important insights:

\begin{itemize}
    \item Both MT2ST strategies outperform standard fine-tuning (STL) on all metrics, indicating that auxiliary guidance helps improve generative fidelity and semantic alignment.
    \item MT2ST-S achieves the best FID and CLIP score, demonstrating better visual quality and text-image consistency. The sharp performance gain around the switching step (400K) supports the benefit of a staged training process.
    \item Reconstruction loss is lower for both MT2ST variants, showing that incorporating auxiliary pixel-level loss early helps stabilize training.
    \item In terms of efficiency, MT2ST-S achieves 31.9\% compression and reduces training time by nearly 19 hours, without sacrificing generative quality.
\end{itemize}

\section{Conclusion}
In this work, we propose MT2ST, a general and adaptive multi-task to single-task training framework designed to accelerate model convergence while preserving or even improving final task performance. MT2ST introduces two complementary strategies—Diminish and Switch—that enable smooth or staged transitions from multi-task sharing to single-task specialization. We evaluate MT2ST across a wide spectrum of models and modalities, including classical representation learners, transformer-based architectures, and diffusion models. Empirical results on text, image, and multimodal tasks show that MT2ST consistently improves accuracy while reducing training time and computational overhead. Our analysis highlights MT2ST as a practical and modular framework for efficient optimization across diverse AI systems.

% \section*{Limitations}
% While MT2ST demonstrates promising results across multiple architectures and modalities, several limitations remain. First, although we validate MT2ST on vision and language tasks, its effectiveness on audio, video, or real-time streaming applications remains to be investigated. Second, current strategies are based on manually scheduled transitions; future work could explore reinforcement learning or meta-learning-based adaptive transition control. Third, our framework still relies on full backbones during training, meaning memory usage remains similar to baseline MTL setups. Finally, as with all multi-task methods, the choice and quality of auxiliary tasks heavily influence the effectiveness of transfer, requiring careful dataset curation and compatibility checks.

\section*{Limitations}
While MT2ST performs consistently well across diverse models and tasks, there still a few aspects can be further refined. Currently, task transition schedules in both strategies are predefined; future work may benefit from more adaptive or learned scheduling. 

\vspace{12pt}

\bibliographystyle{plainnat}
\setlength{\bibsep}{0pt plus 0.3ex}
\bibliography{main}

%\pagebreak
\appendix

\section{Experimental Results Figures}

This section includes the figures corresponding to the experimental results presented in the main text.

\subsection{Single-task Fine-Tuning}

\begin{figure}[H]
    \centering
    \begin{minipage}{0.24\linewidth}
        \includegraphics[width=\linewidth]{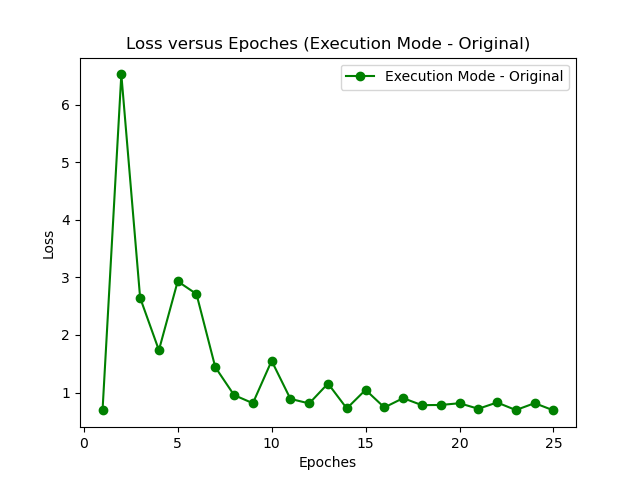}
    \end{minipage}%
    \begin{minipage}{0.24\linewidth}
        \includegraphics[width=\linewidth]{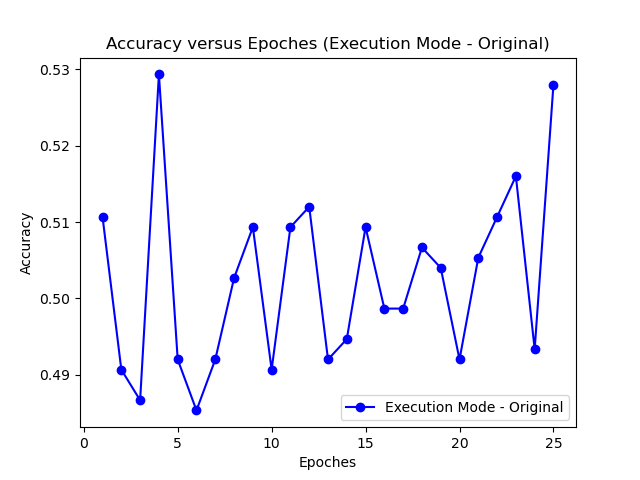}
    \end{minipage}%
    \caption{Loss and Accuracy Change for Single-task Fine-Tuning}
    \label{fig:single_task_loss}
\end{figure}

Figure \ref{fig:single_task_loss} shows the loss and accuracy changes for the single-task fine-tuning approach.

\subsection{Multi-task Learning (MTL)}

\begin{figure}[H]
    \centering
    \begin{minipage}{0.24\linewidth}
        \includegraphics[width=\linewidth]{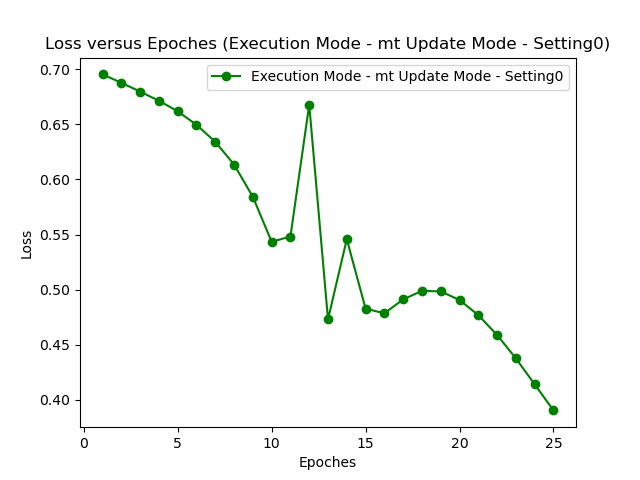}
    \end{minipage}%
    \begin{minipage}{0.24\linewidth}
        \includegraphics[width=\linewidth]{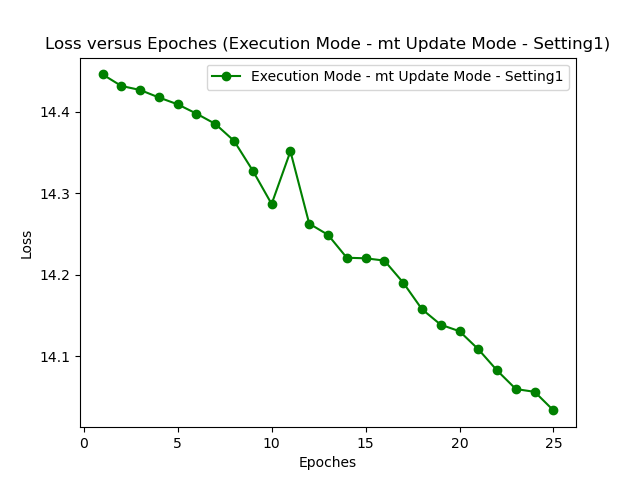}
    \end{minipage}%
    \begin{minipage}{0.24\linewidth}
        \includegraphics[width=\linewidth]{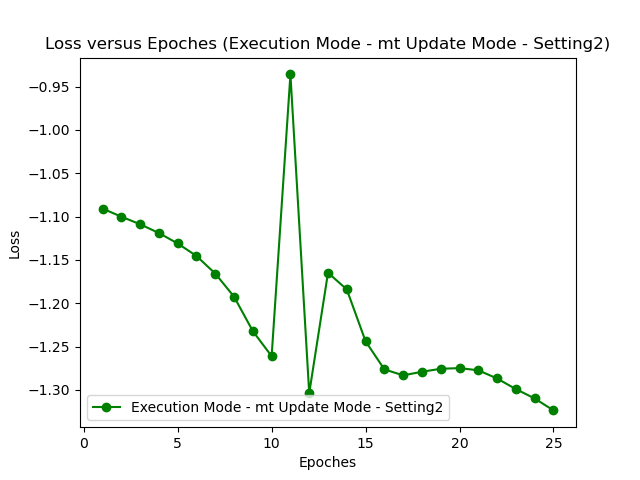}
    \end{minipage}%
    \begin{minipage}{0.24\linewidth}
        \includegraphics[width=\linewidth]{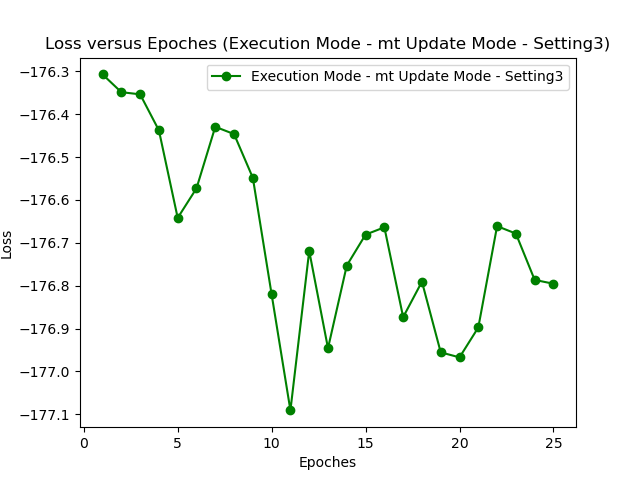}
    \end{minipage}
    \caption{Loss Change in Multi-task Learning}
    \label{fig:mtl_loss}
\end{figure}

\begin{figure}[H]
    \centering
    \begin{minipage}{0.24\linewidth}
        \includegraphics[width=\linewidth]{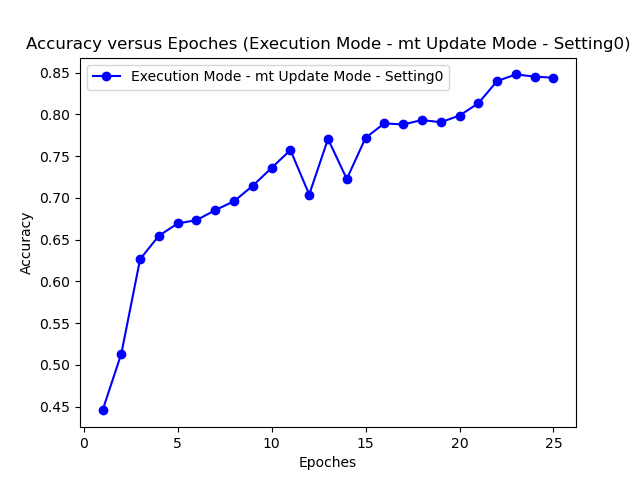}
    \end{minipage}%
    \begin{minipage}{0.24\linewidth}
        \includegraphics[width=\linewidth]{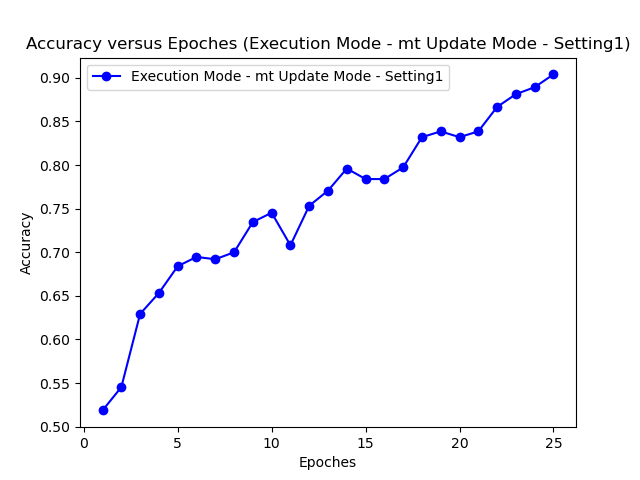}
    \end{minipage}%
    \begin{minipage}{0.24\linewidth}
        \includegraphics[width=\linewidth]{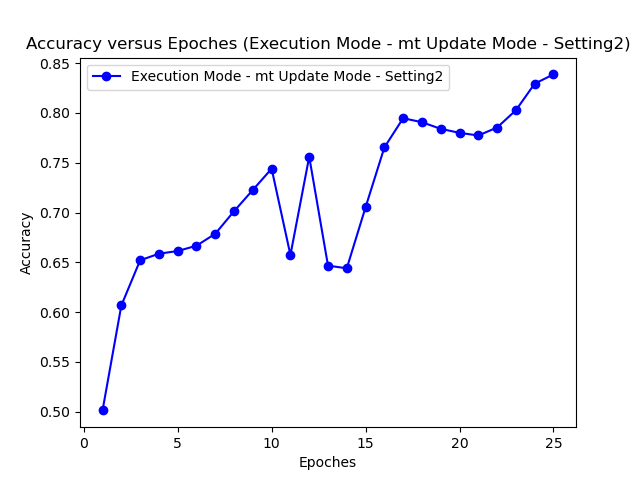}
    \end{minipage}%
    \begin{minipage}{0.24\linewidth}
        \includegraphics[width=\linewidth]{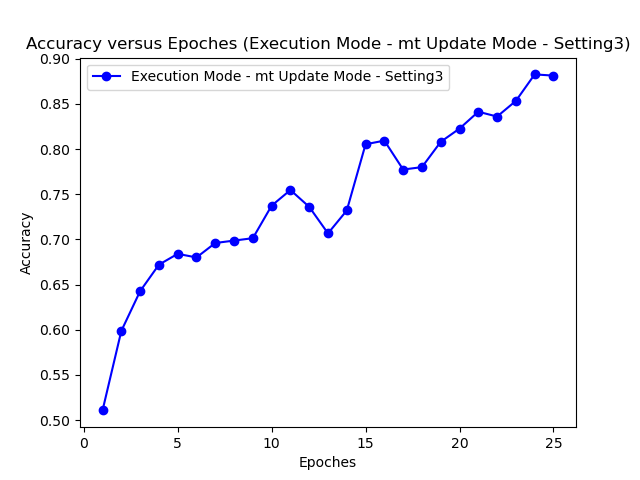}
    \end{minipage}
    \caption{Accuracy Change in Multi-task Learning}
    \label{fig:mtl_accuracy}
\end{figure}

Figures \ref{fig:mtl_loss} and \ref{fig:mtl_accuracy} show the loss and accuracy changes for the multi-task learning approach.

\subsection{MT2ST: Diminish Strategy}

\begin{figure}[H]
    \centering
    \begin{minipage}{0.24\linewidth}
        \includegraphics[width=\linewidth]{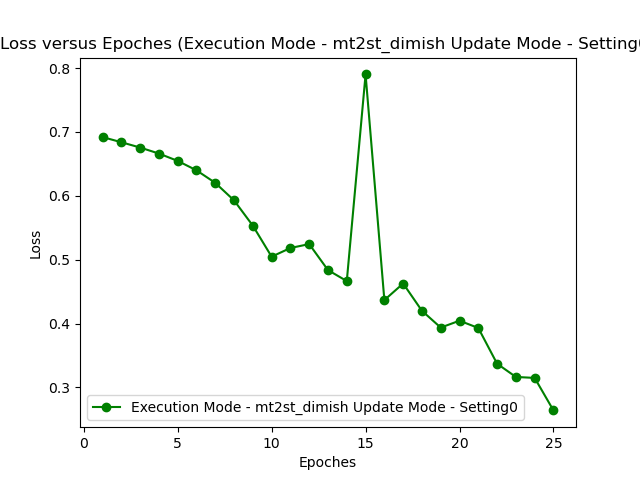}
    \end{minipage}%
    \begin{minipage}{0.24\linewidth}
        \includegraphics[width=\linewidth]{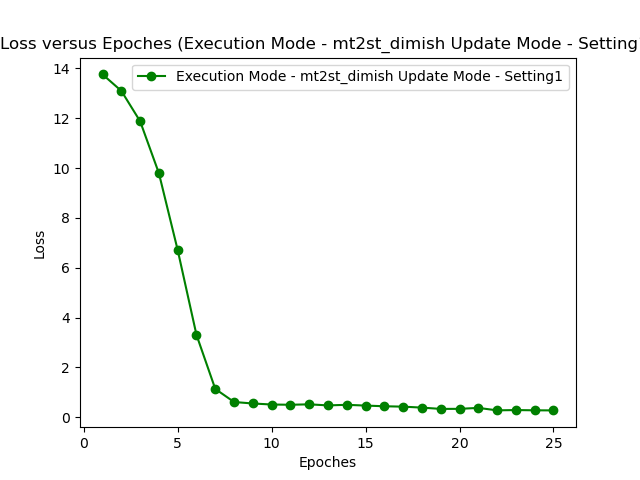}
    \end{minipage}%
    \begin{minipage}{0.24\linewidth}
        \includegraphics[width=\linewidth]{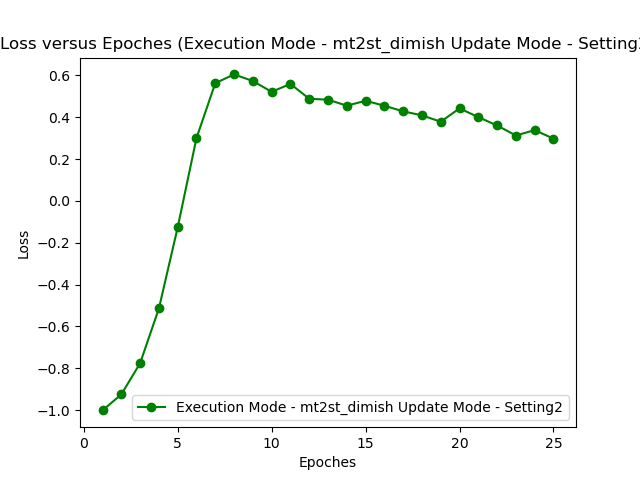}
    \end{minipage}%
    \begin{minipage}{0.24\linewidth}
        \includegraphics[width=\linewidth]{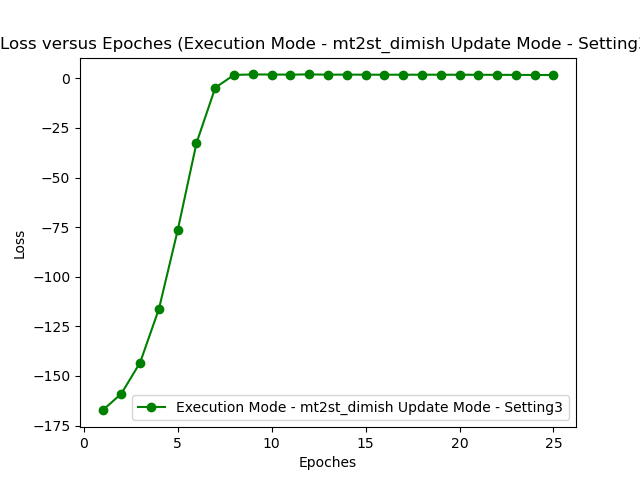}
    \end{minipage}
    \caption{Loss Change in MT2ST: Diminish Strategy}
    \label{fig:mt2st_diminish_loss}
\end{figure}

\begin{figure}[H]
    \centering
    \begin{minipage}{0.24\linewidth}
        \includegraphics[width=\linewidth]{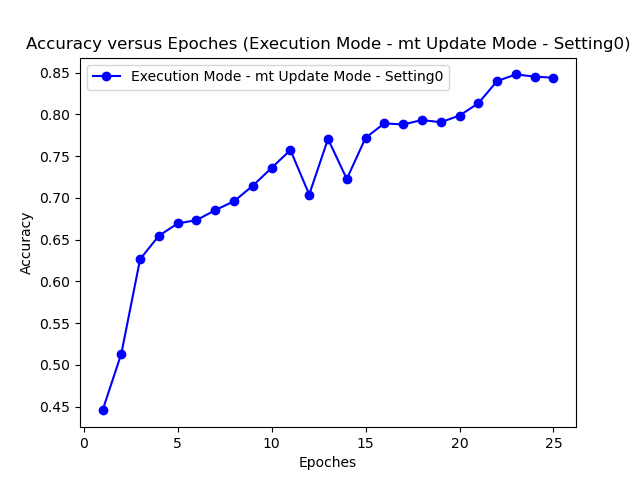}
    \end{minipage}%
    \begin{minipage}{0.24\linewidth}
        \includegraphics[width=\linewidth]{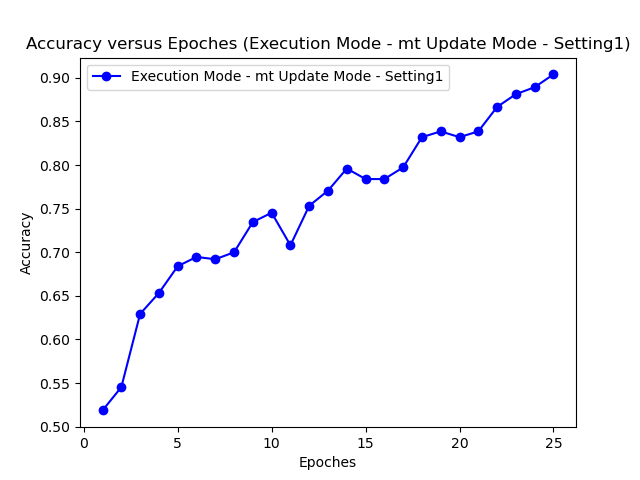}
    \end{minipage}%
    \begin{minipage}{0.24\linewidth}
        \includegraphics[width=\linewidth]{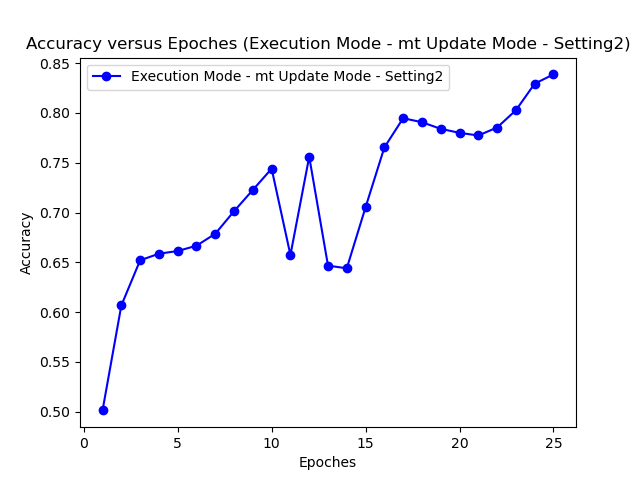}
    \end{minipage}%
    \begin{minipage}{0.24\linewidth}
        \includegraphics[width=\linewidth]{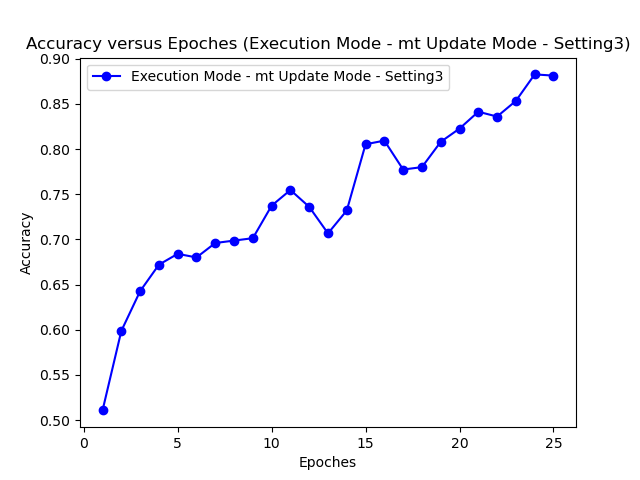}
    \end{minipage}
    \caption{Accuracy Change in MT2ST: Diminish Strategy}
    \label{fig:mt2st_diminish_accuracy}
\end{figure}

Figures \ref{fig:mt2st_diminish_loss} and \ref{fig:mt2st_diminish_accuracy} show the loss and accuracy changes for the MT2ST-diminish strategy.

\subsection{MT2ST: Switch Strategy}

\begin{figure}[H]
    \centering
    \begin{minipage}{0.24\linewidth}
        \includegraphics[width=\linewidth]{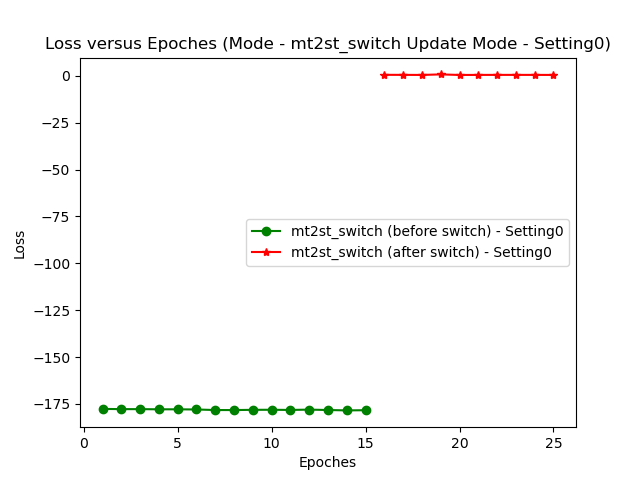}
    \end{minipage}%
    \begin{minipage}{0.24\linewidth}
        \includegraphics[width=\linewidth]{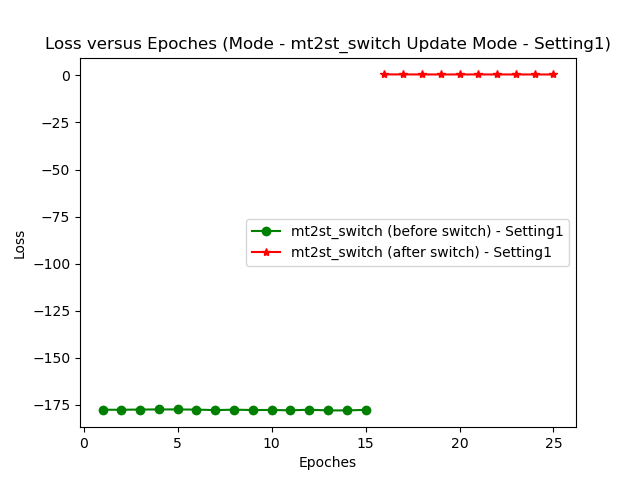}
    \end{minipage}%
    \begin{minipage}{0.24\linewidth}
        \includegraphics[width=\linewidth]{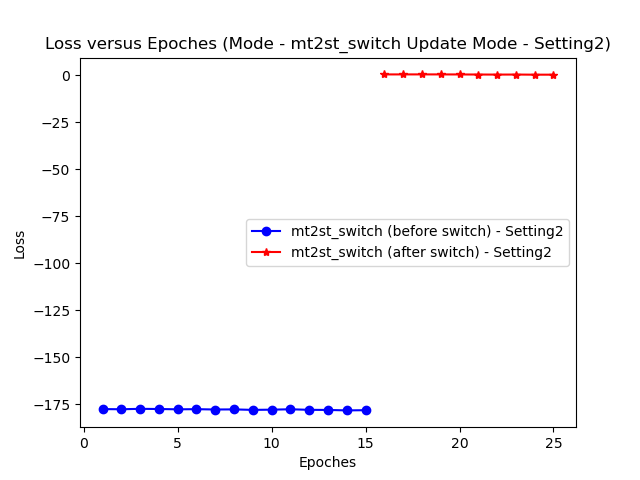}
    \end{minipage}%
    \begin{minipage}{0.24\linewidth}
        \includegraphics[width=\linewidth]{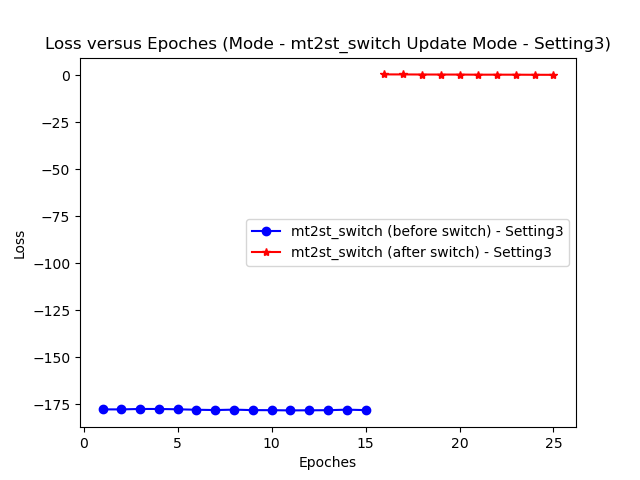}
    \end{minipage}
    \caption{Loss Change in MT2ST: Switch Strategy}
    \label{fig:mt2st_switch_loss}
\end{figure}

\begin{figure}[H]
    \centering
    \begin{minipage}{0.24\linewidth}
        \includegraphics[width=\linewidth]{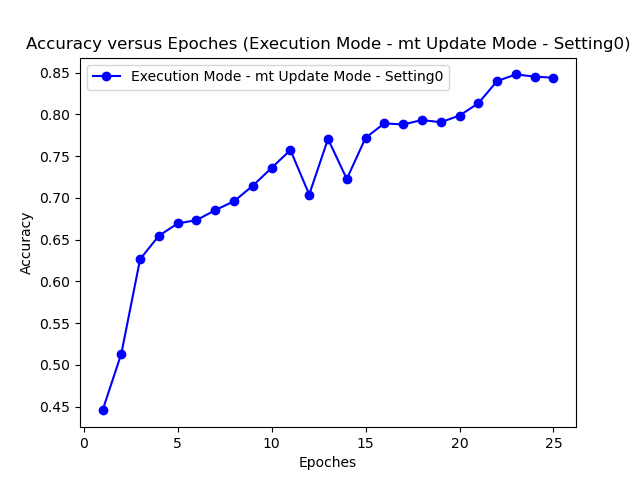}
    \end{minipage}%
    \begin{minipage}{0.24\linewidth}
        \includegraphics[width=\linewidth]{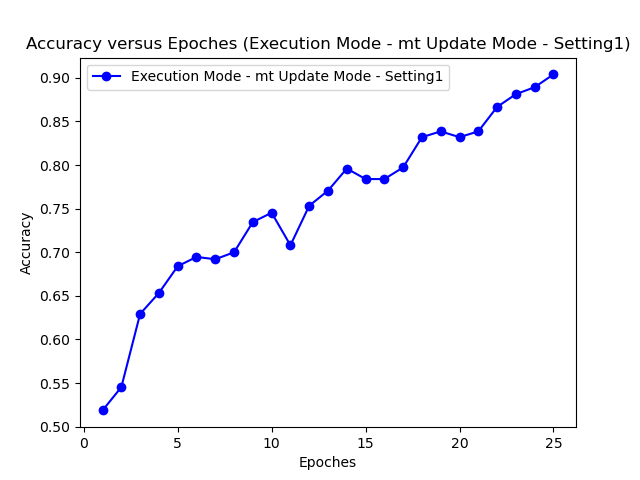}
    \end{minipage}%
    \begin{minipage}{0.24\linewidth}
        \includegraphics[width=\linewidth]{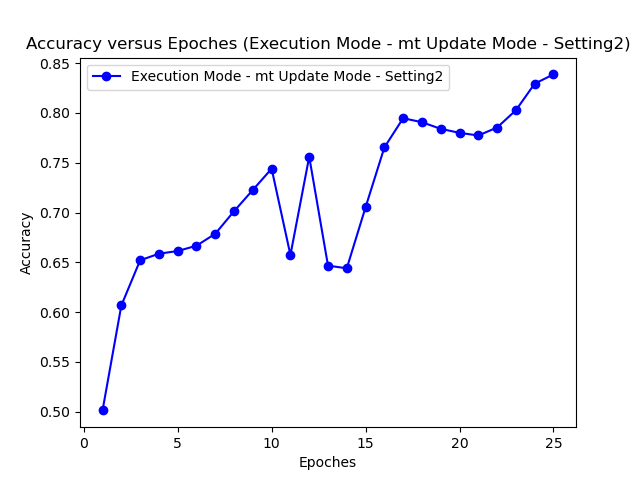}
    \end{minipage}%
    \begin{minipage}{0.24\linewidth}
        \includegraphics[width=\linewidth]{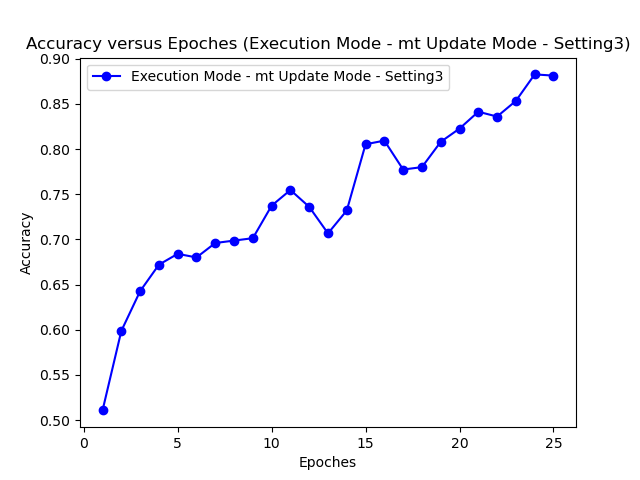}
    \end{minipage}
    \caption{Accuracy Change in MT2ST: Switch Strategy}
    \label{fig:mt2st_switch_accuracy}
\end{figure}

Figures \ref{fig:mt2st_switch_loss} and \ref{fig:mt2st_switch_accuracy} show the loss and accuracy changes for the MT2ST-switch strategy.

\section{Theoretical Foundation of MT2ST}
\label{sec:theory}

In this section, we provide a formal theoretical framework for MT2ST. We first describe a general overview of our method. Then, we instantiate it in the context of shared neural representation learning. Finally, we conduct a theoretical efficiency analysis comparing MT2ST with standard MTL and STL baselines.

\subsection{Overview of MT2ST}

Let a model be denoted by $f(\cdot; \theta)$, trained on a set of $K$ tasks $\{\mathcal{T}_1, \ldots, \mathcal{T}_K\}$. The total loss at step $t$ is a weighted combination of the primary task $\mathcal{T}_{\text{main}}$ and auxiliary tasks:
\begin{align}
\mathcal{L}^{(t)} = \mathcal{L}_{\text{main}}^{(t)} + \sum_{k \neq \text{main}} \gamma_k^{(t)} \mathcal{L}_{k}^{(t)},
\end{align}
where $\gamma_k^{(t)}$ is a time-varying weight for auxiliary task $k$ at iteration $t$. MT2ST alternates between two core strategies:
\begin{itemize}
    \item \textbf{Diminish}: Gradually decreases each $\gamma_k^{(t)}$ to zero over time, enabling soft transition from MTL to STL.
    \item \textbf{Switch}: Explicitly sets $\gamma_k^{(t)} = 0$ after a predefined step $T_{\text{switch}}$, performing a hard switch to STL.
\end{itemize}

\subsection{Formulation of Diminish Strategy}

In the Diminish strategy, each auxiliary task's contribution is governed by a decay function:
\begin{equation}
\gamma_k^{(t)} = \gamma_{k,0} \cdot \exp\left(-\eta_k t^{\nu_k} \right), \quad k \neq \text{main},
\end{equation}
where $\gamma_{k,0}$ is the initial importance of task $k$, $\eta_k$ is the decay rate, and $\nu_k$ controls curvature (decay speed). The overall parameter update is given by:
\begin{equation}
\theta^{(t+1)} = \theta^{(t)} - \alpha \left( \nabla \mathcal{L}_{\text{main}}^{(t)} + \sum_{k \neq \text{main}} \gamma_k^{(t)} \nabla \mathcal{L}_{k}^{(t)} \right),
\end{equation}
where $\alpha$ is the learning rate.

\subsection{Formulation of Switch Strategy}

The Switch strategy introduces a discrete schedule:
\[
\gamma_k^{(t)} = \begin{cases}
1, & t < T_{\text{switch}} \\
0, & t \geq T_{\text{switch}}
\end{cases}
\quad \text{for all } k \neq \text{main}.
\]
The update rule becomes:
\begin{align}
\theta^{(t+1)} = \theta^{(t)} - \alpha \left( \nabla \mathcal{L}_{\text{main}}^{(t)} + \sum_{k \neq \text{main}} \gamma_k^{(t)} \nabla \mathcal{L}_{k}^{(t)} \right),
\end{align}
but reduces to standard single-task learning for $t \geq T_{\text{switch}}$.

\subsection{Theoretical Efficiency Analysis}

We compare MT2ST with baseline MTL and STL methods in terms of convergence behavior and computational efficiency.

\paragraph{Training Cost (FLOPs)} Let $C_{\text{mtl}}$ and $C_{\text{stl}}$ denote per-step FLOPs for MTL and STL respectively. Then, the expected training cost for MT2ST is:
\begin{equation}
C_{\text{MT2ST}} = \sum_{t=1}^{T} \left[ C_{\text{stl}} + \sum_{k \neq \text{main}} \gamma_k^{(t)} C_k \right],
\end{equation}
where $C_k$ is the marginal cost for task $k$. When $\gamma_k^{(t)} \rightarrow 0$ quickly, the training cost approaches STL but retains MTL's benefit in early stages.

\paragraph{Convergence Behavior} Define the effective gradient at step $t$ as:
\[
\nabla \mathcal{L}_{\text{eff}}^{(t)} = \nabla \mathcal{L}_{\text{main}}^{(t)} + \sum_{k \neq \text{main}} \gamma_k^{(t)} \nabla \mathcal{L}_k^{(t)}.
\]
Under the Polyak-Łojasiewicz (PL) condition \cite{karimi2016linear}, MT2ST retains linear convergence rate as long as the auxiliary task gradients align or diminish quickly:
\[
\langle \nabla \mathcal{L}_{\text{main}}^{(t)}, \nabla \mathcal{L}_{\text{eff}}^{(t)} \rangle > 0.
\]
Our strategy ensures that gradient interference is minimized over time, either smoothly (Diminish) or discretely (Switch), avoiding divergence seen in conventional MTL \cite{yu2020gradient}.

\paragraph{Memory Usage} Because MT2ST shares the same encoder across tasks, model memory cost is no worse than MTL. When $\gamma_k^{(t)}=0$, the auxiliary gradients and heads can be dropped from the computation graph entirely.

\end{document}